**Modelling Language using Large Language Models**

**Jumbly Grindrod**
**University of Reading**

Accepted at *philosophical studies*. Please cite published version

**Abstract**

*This paper argues that large language models have a valuable scientific role to play in serving as scientific models of public languages. Linguistic study should not only be concerned with the cognitive processes behind linguistic competence, but also with language understood as an external, social entity. Once this is recognized, the value of large language models as scientific models becomes clear. This paper defends the position against a number of arguments to the effect that language models provide no linguistic insight. Building upon Weisberg's (2007) notion of a model construal, it is then argued that recent work in computational linguistics to better understand the inner workings of large language models can be used to develop a model construal for large language models as models of a language.*

## 1. Introduction

Can large language models (LLMs) inform linguistic inquiry? On this topic we find a divide between those who argue that they can (Baroni 2022; Piantadosi 2023; Westera and Boleda 2019) and those who argue that they cannot (Chomsky, Roberts, and Watumull 2023; Dupre 2021). Perhaps unsurprisingly, the divide largely runs along disciplinary lines, with those on the positive side coming from computational sciences (particularly computational linguistics) and those on the negative side coming from elsewhere.

I will argue that LLMs can shed light on the nature of language. But contrary to those who have previously defended the positive position, I will argue that LLMs do not do so by providing theories of a language. Instead, I will argue that LLMs can fruitfully be thought of as models in the scientific sense: as a structure that serves as a proxy for a real world phenomenon. The real world phenomenon in question is not the cognitive phenomenon of linguistic competence, but the external public language adopted by a linguistic community. Treating LLMs as models in this sense opens up an exciting new branch of linguistic inquiry that was not available to us before the advent of LLMs. I will argue for this claim as follows. I will begin by introducing previous arguments given for the positive view that LLMs can inform linguistic inquiry, before turning to objections against the view. I will respond to those objections by arguing for two key claims. First, LLMs can provide insight into the nature of language understood as an external entity – what Chomsky (1986) has labelled an "E-language". Second, LLMs can do so by serving as models of E-languages. They serve as a representation of an E-language but leave the job of theory-construction still to be done. I will end by responding to the objection that if LLMs are models of anything, then they are really just models of their training data.

## 2. Large language models and E-languages

The idea that LLMs could serve as theories initially looks like a category error. After all, the most famous use of LLMs is as chatbots, producing new text in response to human-inputted prompts. At their core, LLMs will assign probabilities to all words in a vocabulary given some text as input

– often, it is the probability that the word will appear at some point in the input text. This is obviously nowhere near the job that scientific theories are put to. Precisely what scientific theories aim to do and how they do it is itself a controversial topic, but even without a complete account, the following claim is widely-accepted enough for present purposes: in order for a theory to be successful, it must at least have some predictive and explanatory power regarding its target phenomenon. It has predictive power to the extent that it can successfully assign probabilities to an event occurring and it has explanatory power to the extent that it can tell us *why* an event occurs.[1] The category error thought then is that LLMs as systems that assign next word occurrence probabilities are not in the business of providing any kind of explanation and only provide predictions for something that no-one is trying to produce a theory of: which word comes next in any given text.

But this line of thought is too quick. Anyone who has engaged with a LLM chatbot will know that they at least simulate remarkable linguistic ability (as well as a great deal of worldly knowledge), and so the question then becomes whether LLMs have gained this ability by correctly tracking the kinds of linguistic features that are usually captured within a linguistic theory (be they syntactic, semantic, pragmatic, or otherwise). If that is the case, then it may be that predictions and even explanations could be extracted from LLMs regarding those linguistic features.

With regard to prediction, there have been many promising signs in this direction. The following are just a few examples. Manning et al. (2020) found that syntactic dependency relations can be extracted through a linear transformation of contextualised embeddings representing each word of a sentence in context. Nair et al. (2020) found that distance between contextualized embeddings are a good predictor of human judgments of semantic relatedness among polysemous expressions. Shain et al. (2024) found that the next word probabilities that LLMs assign each word serve as a good predictor of human surprisal. Finally, Schrimpf et al. (2020) found that the internal activation patterns of LLMs can be used to predict neural and behavioural responses to texts, in some cases capturing 100% of the variance in their neural response data.

On the basis of the predictive potential suggested by these results, some within computational linguistics are remarkably optimistic about their theoretical potential. For instance, Baroni (2022) has argued that we should view LLMs (which he refers to here as "deep nets") as:

> …linguistic theories, encoding non-trivial structural priors facilitating language acquisition and processing. More precisely, we can think of a deep net architecture, before any language specific training, as a general theory defining a space of possible grammars, and of the same network trained on data from a specific language as a *grammar*, that is, a computational system that, given an input utterance in a language, can predict whether the sequence is acceptable to an idealized speaker of the language. (Baroni 2022, 7)

Similarly, Piantadosi (2023) has argued that LLMs should be viewed as linguistic theories, as they are predictive, precise, and integrate well with related fields such as neuroscience and psychology.

[1] Note that this is distinct, even if related to, the issue of why *we should believe* a particular event will occur or has occurred. For further discussion of the distinction between explanatory facts and evidential facts, see: (Salmon 1989, 6-7).

Indeed, Piantadosi even goes so far as to argue that given the success of LLMs, the generative linguistic tradition as a rival theoretical approach can be rejected.

One initial worry behind treating LLMs as linguistics theories is that LLMs have typically been created within industry, in the interests of companies such as OpenAI, Meta, and Google, rather than in the interest of linguistic inquiry, and that this will affect many aspects of model construction. For instance, OpenAI's GPT series has been constructed according to the interests of OpenAI and their shareholders, who presumably hope to make Chat-GPT the most widely-used chatbot on the market. This in turn affects all aspects of model construction such as selection of training data, the training procedure (including the pre-training procedure and any fine-tuning), the architecture used, and many further parameters besides. All are selected with the company's goals in mind rather than the goal of linguistic inquiry.

But while it is true that LLMs developed in industry often suffer from a commercial bias, it is important not to make too much of this. A great deal of model construction is concerned in the first instance with producing the state-of-the-art for a range of standard evaluation metrics that seem more concerned with testing for language understanding than meeting some commercial goal (such as the GLUE benchmarks to be discussed in §3). But rather than digging in to each design choice for any particular LLM and considering whether it serves commercial interests or not, it is perhaps better for the likes of Piantadosi and Baroni to idealize away from specific models at this point. The thought would be that the LLMs we currently have show us that an LLM could be constructed that could serve as a linguistic theory, provided that it was constructed without any such commercial bias. For example, it would presumably be built without guardrails biasing its word-choice away from particular topics, and would not be fine-tuned in a way that biases it towards producing friendlier or more helpful-sounding text. For the purposes of this paper, I too will be assuming this idealization, considering whether LLMs shorn of their commercial bias could play some role in linguistic inquiry.

Even ignoring the commercial bias of current LLMs, for some optimism that LLMs could act as linguistic theories is still completely unfounded. One reason for thinking this, that both Baroni and Piantadosi consider, is that there would be nothing surprising in the fact that large neural networks can approximate linguistic cognition simply because neural networks of a sufficient size will approximate *any* possible function given enough training. But as both correctly point out, focusing on the general possibility of neural networks approximating any function misses the fact that particular features of the current state-of-the-art architectures have led to this massive increase in progress in recent years. The transition from recurrent neural networks to transformers is a good example of this (Vaswani et al. 2017). The fact that the transformer architecture has proven to be much more capable of capturing contextual relations between datapoints at large scales of training is itself informative about how best to go about dealing with linguistic data.[2] Piantadosi and Baroni then claim that we should expect human linguistic cognition to process linguistic data in a similar manner. So it is not enough to point to the fact that neural networks are "universal function approximators". Even among the universal function approximators, some will be better at processing linguistic data than others, and this, Baroni and Piantadosi suggest, is a basis for linguistic insight.

[2] More specifically, both recurrent networks and transformers are designed to process datapoints in a manner that is sensitive to its surround. Recurrent networks do this through recurrent processing, where a hidden state is used to represent the immediately prior datapoint. Transformers, on the other hand, use self-attention heads that transform how each datapoint is represented based on the surrounding datapoints. The key benefits of the transformer approach is thought to be that data does not have to be processed sequentially and that longer range contextual relations between datapoints are more easily captured.

Other forms of argument do have more bite, however. As has been widely noted, the training that LLMs undergo in order to gain their remarkable abilities is radically unlike human linguistic training. Whatever it is the child is doing, they are not being trained on 500 billion tokens of linguistic data in the way that, for instance, GPT-3 was. The massive discrepancy in training signal alone is reason to think that LLMs do not process language in anything like the way humans do.

When we turn to *how* LLMs learn a language, we also find principled reasons for thinking that they are different to human language users. For instance, any allowance of some form of Universal Grammar thesis (UG) would be sufficient to show that LLMs are quite unlike us (Chomsky, Roberts, and Watumull 2023). The idea of UG is that there are certain innate restrictions on the languages that can be learnt by a human child, and that these restrictions are crucial insofar as they massively reduce the various possibilities that the language learner has to rule out when becoming competent in a particular language. From this, it would follow that there are languages that could be constructed that humans would not be able to learn, for they do not meet the restrictions of UG. But for any such humanly-impossible language, a LLM could be constructed that is trained on data from that language with the same resultant impressive performance that we see in LLMs today. Again, whatever it is the LLMs are doing, it is quite different to what humans are doing.[3]

Finally, it has been argued that LLMs are trained to track patterns of linguistic performance rather than linguistic competence, yet it is the latter that is the proper object of linguistic study. Dupre (2021) has made this case. As he notes, this would not be problematic if performance data was a straightforward outcome of speaker competence. In that case, it wouldn't be implausible to think that there is a way of inferring what competence amounts to on the basis of the patterns in the performance data. However, as Dupre argues, there are several reasons for thinking that there is a wide gap between competence and performance. First, current linguistic theory has good reason to take sets of perfectly ordinary utterances as ungrammatical as far as the internal language faculty is concern. For example, *echo questions*, such as "You're going where?" and *subject-dropping constructions* such as "Had lunch at Franky's earlier" have both been thought to pose problems for accounts of English syntax. While there might be possible grammars that accommodate these constructions, the massive increase in complexity required is good reason to put such constructions to one side as "learned exceptions" to the grammar. Second, Dupre doubts that the hierarchical structures typically posited by linguistic theory could be retrieved by a model that is merely trained to predict the linear progression of text (although here Dupre seems to run afoul of the findings from Manning et al. (2020) discussed earlier). Third, he notes that linguistic theory does not merely seek to capture performance data; in attempting to capture linguistic competence understood as a biological faculty, it will be sensitive to other considerations such as cross-linguistic data, neurobiological evidence, and evolutionary considerations. By contrast, LLMs are essentially trained to replicate performance data in being trained to produce text that is statistically plausible given the training data; biological and evolutionary considerations play absolutely no role in its training.

[3] One kind of response to this argument is that the remarkable learning abilities of LLMs undercuts the motivation for UG. As indicated, one reason for accepting UG is that it is the only solution to the so-called *poverty of the stimulus* problem: that children are able to acquire linguistic competence with use of a relatively small amount of training data. But for some, LLMs show that learning can take place even without prior constraints given as to the outcome (Piantadosi 2023, 18–20). Of course, for this kind of response to work, it would need to be shown that such unconstrained learning could occur with an amount of training data roughly proportionate to a child's. This has spawned the "Baby LM" challenge: https://babylm.github.io/ As this response depends on the outcome of a currently open challenge, I will ignore it for the purposes of this paper.

These arguments build quite a case against the idea that LLMs could serve as theories of linguistic cognition. If we accept these arguments, then the natural conclusion to reach is that LLMs may be a surprising source of predictive power but for the above reasons have little explanatory power. There is something right to this line of thought, and indeed, the view defended in this paper is committed to a version of that claim. But it also overlooks an important form of response, which is to deny that a linguistic theory has to be a theory of linguistic *cognition*. Specifically, the above arguments take a cognitive view of linguistics, that the proper object of study is what Chomsky (1986) calls the *I-language* ("I" for "internal"). Chomsky has repeatedly argued that linguistics is best understood as a branch of psychology and thus biology, and so the linguist should take the internal computational processes required for linguistic competence as their object of study.[4] Or as he sometimes puts it, the object of study for the linguist is the initial state of the language faculty and the manner in which it reaches its final state. If we understand the object of study for linguistics in this way, then the above criticisms have a great deal of plausibility.

But this view is controversial, for it assumes a particular position regarding the ontology of language. Following Santana (2016), we can distinguish between three broad positions on this matter: mental object theories, social object theories and abstract object theories.[5] As we have seen, mental object theories take languages to be a mental entity, possibly understood in terms of the language faculty or in terms of linguistic competence more generally.[6] Social object theories take a language to exist within a community, usually in terms of a set of linguistic conventions (Lewis 1983; Wiggins 1997; Millikan 2005). Finally, abstract object theories take languages to be akin to mathematical or logical objects (Katz 1981; Katz and Postal 1991).[7]

One way of framing this debate is to assume that only one of these positions can be correct. However, I will follow (Santana 2016; Stainton 2011; 2014) in adopting pluralism about language as the object of study for linguistics. Pluralism is the view that there is an abstract object, social object, and mental object, each worthy of linguistic study. I basically endorse the arguments

[4] Closely related to this idea is Chomsky's skepticism of the possibility of a referential semantic that attributes referential properties to linguistic items (Chomsky 2000). See: (Borg 2012; King 2018) for critiques of Chomsky's attack on referential semantics.
[5] Something like this distinction has been appealed to on a number of occasions. (Katz 1981; Katz and Postal 1991) distinguish between nominalists, Platonists, and conceptualists. As Santana (2016) points out, it is worth replacing Platonism with an abstract object view as many of the arguments for this position merely try to show that linguistic objects are akin to mathematical objects, without really engaging in the further question of the ontology of mathematical objects. I also prefer replacing nominalism with a social object position as it more closely aligns with the focus of this paper. In outlining nominalism, Katz and Postal have the American structuralist movement in mind of Sapir, Harris, and Bloomfield, whereas the alternative view of language I am focused on here takes language to be a social object (regardless of whether that social object reduces down to physical linguistic events, mental events, or some combination of the two).
[6] (Hauser, Chomsky, and Fitch 2002) distinguish between narrow and broad construals of the language faculty. The language faculty narrowly construed is the core computational system responsible for generating complex linguistic representations given a finite set of basic representations (i.e. it performs syntactic operations). The language faculty broadly construed includes the narrowly construed faculty as well as a sensory-motor system, responsible for production and detection of language (i.e. speaking and hearing) and the conceptual-intentional system, responsible for the interpretation of the representations.
[7] This approach is given a clear statement in (Lewis 1983) and was arguably adopted by Frege (1948) and Montague (1970).

Santana and Stainton put forward for pluralism, so the reader is directed to their work on the topic, but I will here cover some of the key points they make.

First, as Santana (2016) carefully details, pluralism takes the various sub-fields of linguistics at their word. The idea that linguistics is concerned with the computational properties of the language faculty may be entirely appropriate for certain projects in linguistics, but seems ill-suited to capture what takes place in sociolinguistics, for example, which will focus on how languages, dialects, and idiolects interact with each other and with other social factors. And as Stainton (2014) notes, phonology is an area of study that seems to appeal to a combination of physical and intentional properties alongside auditory properties. The mental conception of language also doesn't fit well with key ideas in philosophy of language, such as meaning externalism. The upshot of externalism is that there are linguistic facts that cannot be determined by examining the psychological properties of a given individual alone.[8] Rather than relegating these important insights outside of the territory of "linguistics proper", pluralism allows them in.

Second, once we distinguish between arguments that x is a worthy object of study, and arguments that x is the *only* worthy object of study, we find that arguments of the second form fail. We will turn to Chomsky's arguments specifically that E-languages (i.e. language as a social object) are not worthy of study shortly, but Santana carefully dismantles various arguments of the second type, raised against all three of the above positions.

Third, Stainton (2011; 2014) has responded to the worry that the pluralist approach is committed to language ultimately existing as some form of curious ontological hybrid: part mental, part abstract, part social. He argues that to the extent that this is true, it will be true of many things:

> Indeed, our world is replete with such hybrid objects: psychocultural kinds (e.g. dining room tables, footwear, bonfires, people, sport fishing, Caribbean cruises, lasagna, the gel pen, eye makeup, ginger ale, champagne, civic unrest, color television, punk rocks, pornography, incest) intellectual artefacts (college diplomas, drivers' licenses, the Canadian dollar, the heliocentric theory of our solar system, abstract expressionism, *Angry Birds*, Microsoft Office, the U.S. Constitution); and institutions (MIT's department of Linguistics and Philosophy, Disneyworld, ethnomusicology, the IBM corporation, Hinduism and Christianity, the NBA, NAFTA). (Stainton 2014)

There is a weak sense in which eye makeup, for instance, is not a purely physical entity, for it possesses a range of cultural properties to do with dress code, beauty, fashion trends etc. To that extent, eye makeup is a hybrid. So if the pluralist ends up being committed to the claim that language is an ontological hybrid, this isn't particularly problematic. For the purposes of this paper, I will argue that one worthwhile project is to conceive of language as a public, social object. It is worth considering the criticisms that Chomsky has levelled against this project.

His most extended attack on E-languages comes in his (Chomsky 1986). There, he argues that the notion of an E-language is the result of a somewhat artificial idealization of the actual linguistic phenomenon of individual speaker competence. As a result, E-languages play no role in linguistic explanation. Instead, explanation of what makes a given sentence well-formed or meaningful will go entirely via a complete account of the possible and actual I-languages of individual speakers:

> Different I-languages will assign statuses differently in each of these and other categories. The notion of E-language has no place in this picture. There is no issue of correctness with regard to E-languages, however characterized, because E-languages are mere artifacts.
> (Chomsky 1986, 26)

[8] For instance, if Putnam (1975) is right then whether someone is speaking English or Twin-English cannot be determined by inspecting their psychological set-up.

There are two parts to this criticism. He claims first that E-languages provide no explanatory value, and second that E-languages are ontologically suspect in that they are delineated in an arbitrary fashion: for instance the distinction between languages, dialects, and idiolects looks to be one that navigates political and cultural lines as much as it does purely linguistic lines. Both parts of Chomsky's criticism are problematic, however. First, we don't have reason to think that the phenomena that would be captured by a theory of I-languages are the only phenomena worth capturing. As Santana states:

> The only way such a claim is true is if we give 'linguistic fact' a narrow, ad hoc definition which excludes, say, facts about communication or language change. Chomsky and his allies do sometimes attempt just such a narrowing, and while they are justified in doing so for their own immediate research program, they have no grounds to suggest that their ad hoc narrowing is binding on the rest of the discipline. (Santana 2016, 512)

Second, the fact that an object is artefactual does not restrict it from being an appropriate object of scientific study. To pick an obvious and well-worn example, economics is a scientific discipline that makes frequent appeal to artefactual objects, where those objects will be often be delineated somewhat arbitrarily. A study into the wage growth patterns in the North of England will have to make a somewhat arbitrary call on what constitutes the North of England. This does not render the study unscientific. As Stainton (2014) notes, it may be that in raising this form of objection, Chomsky is envisioning linguistics as a hard science – or at least harder than what takes place in the social sciences. But it is, I take it, a controversial position that softer sciences must be replaced by harder sciences, and one that would be highly revisionary across many disciplines.

A related concern is that the objects of study are not just artefactual, but vaguely defined. As Chomsky notes: "We speak of Dutch and German as two separate languages, although some dialects of German are very close to dialects that we call "Dutch" and are not mutually intelligible with others that we call "German" (Chomsky 1986, 15). Scientific inquiry requires a precise object of study, so public languages are not appropriate objects of study, or so the concern goes. There are a few different ways to respond here. As Santana (2016, 511) notes, "the existence of borderline cases does not give compelling reason to be skeptical of a categorization scheme. If they did, all vague predicates would pick out entities inaccessible to scientific study." An alternative form of response is to accept Chomsky's argument but instead pursue some form of Carnap-style explication. This is a process where a vague or otherwise problematic ordinary concept is replaced with a concept that is extensionally similar but more sharply-defined.[9] Finally, following Millikan (2005 ch. 2), we might give an account of public language that is more agnostic about the existence of public *languages*. Millikan argues that public language is a set of overlapping, criss-crossing linguistic conventions. Many of these conventions will cluster among particular communities, and to the extent that they do, there is something we can identify as a distinct language. Millikan is doubtful, though, that there is a notion of English, say, that we can use to identify the English-speaking community and in turn give an account of regularities of that community. In this way, she argues, public language exists, even if public languages are more suspect.

[9] For a recent account of the proper method of explication, see: (Pinder 2022)

A further methodological concern that Chomsky briefly raises is that E-languages would be "derivative, more remote from data and from mechanisms than I-languages and the grammars that are theories of I-languages" (Chomsky 1986, 31). But this obviously depends on what exactly we take as data. A common source of data in the generative tradition has been speaker intuition as an indication of speaker competence, and indeed, if this were the only form of data available to the linguist then it would appear that any insight gleaned regarding E-languages would have to go via I-languages. We saw earlier that Dupre argues that the data relied upon in generative linguistics goes beyond speaker intuition, appealing also to developmental considerations and evolutionary considerations. But even so, these appear to be considerations that will speak first to the biological facts underpinning speaker competence before they could be used to theorise about E-languages. In that respect, Chomsky's concern will still hold. But the mistake here is to identify the data that is relevant to the project of capturing speaker competence understood biologically, and then claim that is the only data going. LLMs take a different form of linguistic data in relying upon a huge record of linguistic performance, that is collected into a single corpus on the assumption that there is something that unites it i.e. that it is part of the same public language.

There is perhaps more to say than I have here about Chomsky's attacks on E-languages over the years. A complete account would also tie in Chomsky's skepticism regarding semantics as an area of inquiry. But the above should suffice to show that the key reasons that have been given for rejecting the project of inquiring into E-language are unconvincing. And of course, once that is the case, the possibility looms that LLMs may prove to be useful for that project. Notice that this shift serves as a departure from the likes of Piantadosi and Baroni, both of whom argue that LLMs will prove useful in cognitive linguistics. This continues a longer-standing trend within computational linguistics and particularly the field of distributional semantics (which serves as a theoretical basis for LLM technology) of arguing that the approach could be relevant to cognitive linguistics (e.g. see: (Baroni, Bernardi, and Zamparelli 2014)).

So far we have seen that the arguments that LLMs could not provide linguistic insight fail, for they do not rule out the possibility that LLMs provide insight into E-languages. But what reason do we have for thinking that LLMs do in fact track the features of E-languages?

As stated earlier, E-languages are plausibly thought of as a set of linguistic conventions, and so arguably the most detailed inquiry into the nature of E-languages has come from the Lewisian (1969; 1983) project of providing a theory of conventions and linguistic conventions in particular. Lewis defended an elaborate theory of conventions as regularities that are adopted as solutions to coordination problems, which included conditions on the beliefs and preferences of those within the community. Regarding language specifically, he argued that a linguistic community adopts a convention to use a specific language when they try to be truthful and trustworthy in performance and audition of sentences in that language. But while his theory is still the starting point for much discussion of the topic, nearly all parts of his theory have proven controversial. Some have challenged whether conventions are really regularities (Millikan 2005);

others have challenged the intellectual requirements of the theory (Burge 1975); others still have criticized Lewis for failing to capture the normative nature of conventions (Marmor 1996).

As important as these debates are, I don't need to make a call here on every aspect of the nature of conventions. Instead, to make a case that LLMs do track features of languages understood conventionally, I just need to appeal to claims that all who adopt a conventionalist position would agree with. First, whatever it is linguistic conventions are, they exist at the group level rather than the individual. Second, there is a direct dependence between these conventions existing and actual language use: that a convention has been adopted must be reflected in actual usage. Third, included within the set of linguistic conventions are the full range of linguistic properties that one finds investigated across linguistics, including syntactic, semantic, phonological etc.

Given that conventions exist at the group level and depend for their adoption on actually being used, the best chance we have of tracking those conventions empirically will be by working with data that tracks linguistic activity across a large group of people. This is precisely what LLMs are trained on. They are trained on a huge collection of linguistic data that records repeated usage of linguistic items over time. As such, the patterns that persist in the data and that will be tracked by a successful LLM will include linguistic conventions. Clearly, LLMs, in utilizing all regularities in the training data that will prove beneficial to their language prediction task, pick up on more than just linguistic conventions. This is most clearly evidenced by the impressive amount of factual information that they become sensitive to. But it is enough for present purposes that included within what LLMs do track are the patterns of behaviour that conventionalists about language would identify as conventions. This gives us at least initial reason to think that LLMs are well-suited to track E-languages. And initial reason is really all that is required to motivate an empirical project, where ultimately the proof will be in the pudding.

## 3. Large language models as models of languages

In the previous section, I have argued that the arguments to the effect that LLMs could not provide linguistic insight fail provided that we take them as tracking E-languages, and, furthermore, that there is initial reason for thinking that they do just that. Let us turn then to the second key claim of the paper, that LLMs are best thought of as models of E-languages rather than as theories. To make good of this claim, we first have to be clear on what exactly we mean by a model, and we then have to consider whether LLMs could fruitfully serve as a model in our favoured sense.

Initially, the idea of a scientific model doesn't seem particularly difficult to grasp. Sometimes, instead of obtaining data on the target phenomenon, scientific inquiry will proceed by constructing an object that serves as a proxy for the target phenomenon and then investigating that.[10] The object in question serves as a proxy insofar as it is known to share a number of relevant properties with the target phenomenon, and so it is reasonable to expect that it will behave in the same way as far as the inquiry is concerned. To use Hesse's (1970) analogical terminology, the properties that are known to be shared between the model and the target are the *positive analogy*; the properties that are known not to be shared are the *negative analogy*; and the properties for which we do not know whether they are shared are the *neutral analogy*. One way of viewing the use of models in this kind of inquiry is as an exploration of the positive analogy (and possibly also a minimization of the neutral analogy).

[10] "Object" here needs to be understood very broadly so that it at least includes abstract objects (such as mathematical objects) and computational systems.

However, there are reasons for thinking that the distinction between theory and model is not as clear-cut as it first appears. One of the main shifts in 20th century philosophy of science was an increasing focus on the importance of models to scientific inquiry. Quite apart from being an optional way of investigating the target phenomenon, it has been argued on a number of occasions that models play an essential scientific role. Perhaps most famously, it has been argued that scientific theories are ultimately to be understood in terms of models. The shift by the likes of Van Fraassen (1980; 1989) and Suppes (2002) away from the syntactic view of scientific theories and towards the semantic view is one that had models at it heart.[11] Whereas the syntactic view claims that a theory consists in a set of theoretical sentences, the semantic view claims that a theory consists in a set of models that represent the target phenomenon and that are described by the theoretical sentences. This is partly why there has been something of a modelling turn in philosophy of science, as the role of the model in theorizing has been seen as increasingly important.

Even aside from this increased focus on the role of models in theorizing, there are other reasons for thinking that the model/theory distinction is perhaps not clear-cut. One is that sometimes theories will be used as models. For example, the Bohr model of the atom is a now disproven scientific theory of atomic structure. Nevertheless, its simplicity combined with its relative accuracy regarding particular phenomena means that in particular contexts (including investigative and pedagogical) it is a particularly useful model to employ. A further reason for thinking that the model/theory distinction is not particularly clear-cut is that, as documented well by (Bailer-Jones 2013), the distinction does not appear to be used consistently by practicing scientists.

However, we can allow that models may end up being essential to understanding the structure of scientific theories, and that models may interact with scientific theories in other ways as well, while insisting that there is a distinctive use of models in scientific inquiry such that the claim that LLMs could serve as models in this sense is importantly distinct from the claim that the LLMs could serve as theories. Following Weisberg (2007; 2013), we can distinguish between two strategies of theorizing. On the *abstract direct representation* strategy, observational data are collected directly about the target phenomenon. Decisions have to be made about what properties of the target will be recorded and which are ignored, and the target phenomenon is to that extent abstracted. But as the name suggests, this strategy nevertheless proceeds through an attempt to theorize directly about the target as it is represented in the available data. By contrast, on the *modelling* strategy, a theoretical investigation proceeds through the construction of a model that is sufficiently similar to the target phenomenon in the relevant respects. Weisberg uses the example of the Lotka-Volterra equations that initially served as a model of the predator-prey population levels in the Adriatic sea after the First World War. I suggest that LLMs can be thought of as models of an E-language in this sense.[12]

[11] Here models are understood almost exclusively as mathematical objects. Suppes appealed to models in the model-theoretic sense (i.e. an assignment of values that would make the theoretical sentences true) while Van Fraassen understands models as state spaces where dimensions correspond to theoretical parameters.

[12] Models can fruitfully be used in scientific practice without being taken to represent their target phenomenon. In the philosophical literature, there is considerable discussion of toy models - that are highly simplified or otherwise distorted representations - and minimal models - that arguably do not represent any particular target (Batterman and

Weisberg goes on to detail how model construction consists in two parts: development of a structure (i.e. the object that will serve as the model) and development of an interpretation, or what Weisberg we calls a *construal*. These two parts do not always occur in order or even sequentially – sometimes the construal is modified as the structure is modified. Although construal development is often left implicit, it is nevertheless a crucial part of model construction, as the same structure can be used to represent various different targets, or can represent the same target in a variety of different ways. As an example of the former, Weisberg (2013, 77–78) points to Goodwin's (1967) use of the Lotka-Volterra equations to capture the relationship between employment rates and the workers' share of the national income. A less extreme example lies in the use of the same equations to capture any predator-prey relations beyond the Adriatic sea post-World War I. But even if the target phenomenon remains fixed, it is still up to the modeller to decide what exactly of the target phenomenon is captured by the model and to what degree of accuracy. To capture this, Weisberg identifies four parts to the construal. The *assignment* specifies the parts of the model that represent parts of the target phenomenon. The *scope* specifies which parts of the target phenomenon are captured by the model and which are ignored. There are then two forms of fidelity criteria, both concerned with how accurately the model represents the target. *Dynamical fidelity criteria* "tell us how close the output of the model must be to the output of the real-world phenomenon" (Weisberg 2007, 221). *Representational fidelity criteria* provide a standard for whether the outputs for the model are produced in the same way that the real world outputs are produced, where this is usually understood in terms of some similarity or isomorphism in internal structure.

The need for a construal highlights that the central claim of this paper – that LLMs could serve as models of public languages – is consistent with a wide array of outcomes depending on the precise nature of the construal. As I will argue later, settling on how exactly LLMs should be construed as models should really be viewed as an on-going task. But I am not here endorsing one possibility at the extreme, minimal end of possible construals. We could state that the assignment is that the input-output patterns of the LLM represent production patterns of a language, the scope is just the production of linguistic text, and that there is no mapping of the internal structure of the LLM and some feature of the language. As a result, the construal would only allow for dynamical fidelity criteria and no representational fidelity criteria.[13] In passing Nefdt (2023, 884) suggests that artificial neural networks generally should be viewed in this way: "…contemporary artificial neural network models are remarkably dynamically faithful. What they lack, as is evidenced by the explainable AI movement, is representational fidelity…". While it is certainly true that we can treat LLMs in this way, this is to embark on the project that I earlier suggested no-one is trying to theorize about – that of predicting language production. What I am suggesting instead is that we can take internal features of LLMs to represent the linguistic conventions of a language that regulate linguistic performance. This is obviously a more ambitious project that has much more explanatory potential by capturing why language is produced in the manner that it is, rather than just predicting future usage.

But here an objection looms. For in the case of LLMs and in the way that they are constructed, it doesn't seem possible to develop a construal that appeals in its assignment to internal features of

---

Rice 2014; Reutlinger, Hangleiter, and Hartmann 2018). As a reviewer notes, LLMs could plausibly serve in these roles. For instance, investigating why LLMs struggle to deal with negations (Truong et al. 2023) may provide a basis for studying the phenomenon of negation itself. In this paper, I am concerned with the possibility of using LLMs as representations of target phenomena, but the possibility that LLMs could serve as models that do not represent a target is well worth exploring elsewhere.

[13] Thanks to an anonymous reviewer for raising this possibility.

an LLM. It may be the case (the objection runs) that LLMs do end up tracking all kinds of interesting linguistic features, but this is something that could only be discovered after construction of the model and cannot really inform any modifications of the model. LLMs are constructed using a form of self-supervised learning on a language prediction task. In slightly more detail, given a large training corpus, the LLM is set the task of taking extracts from that corpus, masking one or more words in the extract, and predicting the words that has been masked. It can then check how far away its predictions were from the correct answer using a loss function. Through the use of backpropagation together with an optimization algorithm, the calculated loss can then be used to adjust the weights of all connections across the network so that its predictions better approximate the correct answer. While there are a number of parameters that can be controlled in setting this task (e.g. the nature of the task, the size of the extract, number of training epochs, loss and optimization algorithms etc. ) as well as choices to be made about the architecture and the training data, there is very little direct control over the model that we end up with after training. And of course, the model cannot be modified after training according to the modeler's construal, because LLMs are largely opaque, meaning that a modeller can't know how to modify the inner workings of an LLM so as to better meet the desired construal. As such, it seems that any construction and refinement of the model according to the modeller's construal simply doesn't seem possible here, or so the objection goes. Note that this worry would apply to all parts of the construal: it would not be clear which parts of the LLM are supposed to capture the target phenomenon, what of the target phenomenon would be represented by the LLM, or what the fidelity criteria for the model would look like.

In response to this objection, it should be acknowledged that if LLMs are to serve as models, that there is substantial work to be done to find out how LLMs could serve as representations of the target phenomenon i.e. an E-language. As the above objection makes clear, the issue is partly one of opacity and partly one of control. More specifically, as has been well-discussed regarding deep neural networks more generally, the manner in which they are constructed is so complex – involving thousands or millions or billions of parameters – that understanding how each parameter contributes towards the network's behaviour is simply beyond our ken.[14] Partly because of this opacity, we are not able to directly control the way in which such models are generated beyond the specification of training and architecture mentioned earlier.

There has been some recent discussion in philosophy of science of whether deep neural networks could be a source of explanatory value despite their opacity. Sullivan (2022) has argued that opacity of some algorithmic step that the neural network completes is not a particular barrier to it serving as a basis for understanding.[15] Algorithmic steps can often be broken down into sub-steps. But it is not the case that in order for understanding to be generated off the back of that algorithm, it must thereby be understood how each step and sub-step was implemented. Understanding may be generated merely by understanding how the algorithm works at a higher level without drilling down into each lower algorithmic level, and indeed, sometimes the particular implementational details at the lower levels seem irrelevant to understanding. Shech and Tamir (2023, 329) give the nice example of an underground map serving as proxy for real world geography in such a way that certain implementation details (e.g. the particular colours used for each line, the font of any labels etc.) are irrelevant to the explanatory value of the map.

[14] As Søgaard (2024) details, the source of the opacity of deep neural networks has been put down to a number of features, including their size, non-linearity, the use of continuous variables, their lack of groundedness, and the incrementality of their training. However, I will follow Søgaard in taking it that as far as *inference opacity* (i.e. opacity of why a system produced the output that it did) is concerned, the size of the network is the sole cause for opacity.

[15] For the purposes of the discussion here I will follow Sullivan in assuming that explanation aims at understanding.

The same will be true of models, deep neural networks or otherwise: whether or not certain implementational details are opaque is not of importance.

Rather than just any form of opacity acting as a barrier to understanding, Sullivan argues that the explanatory value of deep neural networks is limited by the level of *link uncertainty* i.e. "the extent to which the model fails to be empirically supported and adequately linked to the target phenomena" (2022, 110). Link uncertainty obviously speaks to the problem of constructing a construal in Weisberg's sense in that to the extent that there is link uncertainty, we will not be able to construct an assignment, scope, or fidelity conditions for the model. So how can link uncertainty be reduced? The main way that Sullivan identifies is by attaining "additional scientific evidence that supports the connection between the causes or dependencies that the model uncovers to those causes or dependencies operating in the target phenomenon" (Sullivan 2022, 113). Sullivan's example of a deep neural network model that achieves this is one that has been trained to identify cases of melanoma from annotated images of melanomic and non-melanomic moles. Here, she argues that the link uncertainty is reduced because there is already extensive scientific backing for the link between melanomas and the surface appearance of moles. As a result, the outputs of that model, as well as its internal workings, is a source of great explanatory value regarding that subject.

Of course, one important response to the issue of opacity has been the suite of techniques that have come out of the *explainable AI* (or *XAI*) research programme. As Zednik and Boelsen (2022, 226) point out, Sullivan (2022, 122) hints that use of XAI strategies could help further reduce link uncertainty by allowing us to check whether the model is processing data in a way one would expect given what we know about the domain. This is particularly important when it comes to LLMs, as in recent years there has been a huge amount of research on producing XAI techniques for LLMs, which take a number of forms. The work from Manning et al. (2020) mentioned earlier is one excellent example, where it was shown that the embedding space within which words are represented in a transformer architecture can be used to extract syntactic dependencies between words in a sentence. But there has also been a great deal of work to show how various linguistic features are processed at various stages in the model (Rogers, Kovaleva, and Rumshisky 2021). One of the distinctive structures within the transformer architecture are the layers of self-attention heads, where each self-attention head will perform a transformation on the vectors for each input word given the words that surround it. As a result, there has been a large amount of work to visualise what each self-attention head does in terms of the kinds of syntactic, semantic, or otherwise linguistic relations that it appears sensitive to.[16] There has also been a great deal of work using *probing classifiers*, where a supervised classifier is trained on the word embeddings outputted by the model or by parts of the model (e.g. particular self-attention heads) to become sensitive to particular distinctions. For example, a probing classifier could be trained on the word embeddings that an LLM assigns each word to see whether it is sensitive to parts-of-speech (POS) categories.[17]

XAI techniques such as these not only reduce the opacity of LLMs, they provide opportunity to consider whether LLMs function as we would expect a model of an E-language to. So the fact that, for instance, LLMs are assigning syntactic dependency relations across words provides good initial reason to think that they can serve as models of a language simply because we would expect to see syntactic relations represented within a model of an E-language.

[16] For instance, see: (Z. J. Wang, Turko, and Chau 2021).
[17] In fact, LLMs are now widely-used to automatically annotate corpora with POS labels.

While XAI techniques help counteract the issue of opacity, the evaluation tasks that LLMs are typically tested against help counteract the issue of control. While many focus on the nature of the training task that LLMs are directly programmed to follow in either pre-training or fine-tuning, the fact that language models are usually evaluated against tasks that more directly test for language comprehension (in some sense of the term) is often overlooked.[18] For example, the GLUE multi-task benchmark (A. Wang et al. 2018) is specifically designed to test models on a number of tasks that together suggest that LLMs have gained general linguistic capabilities.[19] This includes tasks on grammaticality, paraphrase detection, entailment and contradiction detection, and pronoun resolution. So while the proximate training task of each individual LLM may be language prediction, taking a broader perspective on the way that LLMs are constructed reveals an iterative process where LLMs are constructed and then evaluated against tasks designed to test for substantive linguistic features, with their performance then being used to inform future LLM construction. With the use of such evaluation tasks, there is significant, if still indirect, control over the way that LLMs perform. As one small example of this, consider the construction of RoBERTa from BERT. BERT is a transformer LLM initially released in 2019 that has been widely-used and investigated (Devlin et al. 2019). Rather than just being pre-trained on a single language prediction task, BERT is pre-trained on a language prediction task and a sentence continuation task – given two sentences, the task is to give a verdict on whether or not one immediately followed the other. But in an important study into the optimal configuration of BERT, it was found that the sentence comprehension task actually hurts performance on a range of tasks including GLUE tasks (Liu et al. 2019). Liu et al. labelled their subsequent configuration of BERT *RoBERTa* (*Robustly-Optimized BERT Approach*), which has since become a widely-used model. In this way the goal of improving performance on tasks like those in GLUE continues to inform model construction, including even the training task that the model is trained on.

How do we get from these considerations to something like a construal for the model? Let's take syntactic dependencies as one example and consider how we might develop an assignment. Suppose we construct a LLM that performs at state-of-the-art level on the Corpus of Linguistic Acceptability task that is one part of GLUE. The task is to train a classifier that, given input representations of whole sentences (which can be outputted by LLMs), can classify them as acceptable or unacceptable. That our hypothetical LLM performs well on this task is strongly suggestive of the fact that the LLM has learnt to encode those relations that words in a sentence bear with one another and that contribute to the acceptability of the overall sentence i.e. syntactic relations. The earlier findings from Manning et al. (2020) take us a step further in this regard in showing that syntactic dependencies can actually be extracted from the ways that each individual word is represented relative to one another. So that gives us very strong evidence that the model is capturing syntactic relations between words. But in order to generate an assignment, we would ultimately want to go further and state where in the model this takes place. Current work in this direction is suggestive, but following others, suppose that we can then use probing classifiers and similar XAI techniques to identify specific self-attention heads that are sensitive to specific syntactic relations. It will likely not end up being as simple as attributing specific linguistic tasks to specific self-attention heads. Rather each syntactic relation will likely be processed in a distributed fashion, in patterns of activation across various parts of the architecture. But we might still identify particular regions of the architecture – and ultimately particular patterns of activation across those regions – that are responsible for particular syntactic relations. Regarding syntactic relations more generally, Rogers, Kovaleva, and Rumshisky (2021, 847) note that there is a consensus that in BERT models, self-attention heads are sensitive to syntactic features in the earlier layers of the model. Over time, these claims will become more specific, and so we will increasingly have the ability to create a form of assignment regarding syntactic relations.

[18] Pre-training is the more significant training procedure for large language models, which usually uses some language prediction task. This training typically utilizes very large training sets and a great deal of computational power. The resulting language model can then be fine-tuned for more specific tasks. Fine-tuning can take a number of forms, depending on the nature of the more specific task.

[19] The GLUE benchmark has since been succeeded by the SuperGLUE benchmark (A. Wang et al. 2020). However, for purposes of discussion I will focus on GLUE here.

This illustration shows how we can start to develop a construal for LLMs as models of an E-language, with the process described here repeated for other linguistic features and for other parts of the construal. Whereas for simpler models (like the Lotka-Volterra equations) the construal can be relatively straightforward to develop, the development of the construal here is a large exploratory project that utilizes not only what is known about how best to construct models so as to meet evaluation tasks, but also the best XAI techniques we have available to explore how exactly LLMs are able to process linguistic data. But while the construction of the construal may be more laborious than in standard cases of modelling, there is clear potential for a great scientific payoff here. Because LLMs learn to track linguistic features in an autonomous fashion, without prior specification as to what should be captured, we have the opportunity to construct models that capture linguistic features that we wouldn't have otherwise known how to capture.

It is important to note as well that, like the current development of LLMs more generally, construal construction will prove to be a continual and iterative process as model architectures develop and our understanding of those architectures improve. This will include a greater understanding of the particular features of a model that are *not* important parts of the construal. For example, if multiple LLMs are able to achieve a similar, state-of-the-art performance on some evaluation task, and seemingly do so through the same procedure, despite the fact that they diverge on some structural feature, this would suggest that a specification of that structural feature will not figure in the construal (Shech and Tamir 2023, 331). Conversely, and to return to our hypothetical scenario, if it is found that only LLMs that possess a particular structural feature are able to capture our syntactic relation of interest, then this is suggestive that the structural feature will play a role in the construal. And of course, it would also suggest that the feature should be retained in any future model designed to capture syntactic relations. In this way, continual evaluation and construction of LLMs will lead to a continual refinement of a suitable construal of the LLM as a model.

## 4. Objection: LLMs are models of the corpus

Before concluding, I would like to consider an objection that I think naturally arises, and that has been voiced (usually somewhat indirectly) on a number of occasions. This is that if LLMs are to serve as models of anything, they are really just models of their training data.[20] This thought is tempting, particularly when focusing again on the language prediction task that is usually used in pre-training. In training a model to predict unseen text given seen text, we are training a model to be able to reproduce the patterns that it sees in the original data. At its limit, a model that does this perfectly will, given the start of its training corpus as input, be able to predict the rest of the training corpus as output. In this way, LLMs really just provide a way of exploring the original training data, rather than something like an E-language. LLMs are really just "corpus models" as Veres (2022) puts it.

[20] In the field of distributional semantics, Sahlgren (2008, 49) has argued that the distributional models (which plausibly include LLMs) are "models of word meanings… not the meanings that are out there in the world, but the meanings that are in the text". Echoing some of the Chomskyan arguments considered earlier, Veres (2022) has argued that LLMs are best thought of as "corpus models" rather than models of a language. Chiang (2023) has argued that LLMs are best thought of as compressions of their training data, in the same way that jpegs are compressions of higher resolution image files.

This kind of view will run into difficulties, however, given further details of how LLMs are trained. For instance, across machine learning training data is usually split into a training set and a (smaller) validation set. Training consists in the system updating its parameters according to its performance against the training set, but it is then also regularly tested against the validation set, and it is usually the system's performance against the validation set that is taken as a measure of whether training is complete or not. The point of this is to avoid *overfitting* to the training data. If performance against the validation set starts to become worse, this is a sign of overfitting. The type of model that the objection here imagines as an ideal – one that perfectly replicates its training data – would really just be the perfect case of overfitting, and this is just viewed as a standard problem to mitigate against in machine learning. Other techniques are used as well., such as *regularization techniques*. For instance, GPT-3 was trained using a regularization technique known as "weight-decay" (Brown et al. 2020, 43) which will discourage very strong neural weights, and in so doing prevent the model from becoming overly-complex, fitted to all possible features of the original training data.[21]

One way of viewing the matter is that any E-language generates its own ultimate corpus – all possible texts that could be produced in that language. The model is being set the task of predicting what would occur within the ultimate corpus given its exposure to only a small subset of that corpus – the training data. Through techniques such as those mentioned above, the model is guided away from producing predictions by just memorizing the training data, and so is forced to instead generalize based on the patterns it is exposed to. The claim of this paper is that in generalizing on the basis of its training data, the model becomes sensitive to the underlying rules that the training data adheres to and that (provided that the training data is representative) the ultimate corpus will adhere to as well. Included within these underlying rules will be the linguistic conventions that the language consists in.

Further support for this idea comes from taking a broader perspective and considering again the issue of evaluation tasks. As we have seen, although LLMs are constructed using a language prediction task, their performance in that task is not the ultimate desideratum of success. Instead, LLMs are constructed and evaluated based on their performance against a broad banner of language comprehension tasks such as those included with GLUE. The fact that LLMs have made such progress against these benchmarks only further supports the idea that LLMs are not merely memorizing their training data, but are internalizing the rules according to which the training data is produced.

The alternative view of LLMs as models of their training data fails then, as it places too much emphasis on the proximate pre-training task and so relies on a simplistic view of how such models are constructed. LLM construction already employs a range of techniques, in training, fine-tuning, and evaluation, that avoid a mere compressed replication of the training data.

[21] In this respect, LLMs are importantly disanalogous to compression files such as jpegs, *pace* (Chiang 2023). Jpegs in a sense do attempt to overfit to the initial data whereas LLMs and other machine learning models attempt to generalize from the patterns found in the initial data.

## 5. Conclusion

In this paper I have argued that there is good reason to treat LLMs as scientific models for E-languages. I have sought to show this by first responding to arguments to the effect that LLMs could not provide any form of linguistic insight and second by detailing how LLMs could be used as part of a modelling approach. While the use of LLMs as models does bring with it difficulties not found with simpler models, this just means that the process of developing a construal for the model expands into a preliminary exploratory project. But I have sought to show how this project can be completed and given reason to think that it is already underway. In one respect then, the aim of this paper is ambitious: to outline a new kind of linguistic inquiry that LLMs can play a crucial role in as they act as models for E-languages. This should give us cause for optimism for the new forms of insight that these models could provide. In another respect though, the aim of this paper is realistic: engaging in this kind of inquiry in the first instance really is just a matter of reinterpreting the important work that is already being done across computational linguistics to better control and understand LLMs such that they can be a source of scientific insight.